\documentclass[sigconf]{acmart}

\usepackage{bbding}
\usepackage{amsmath}
\usepackage{bbm}
\usepackage{mathtools}
\usepackage{amsfonts}
\usepackage{balance}
\DeclareMathOperator*{\argmax}{arg\,max}

\AtBeginDocument{%
  \providecommand\BibTeX{{%
    \normalfont B\kern-0.5em{\scshape i\kern-0.25em b}\kern-0.8em\TeX}}}

\copyrightyear{2022} 
\acmYear{2022} 
\setcopyright{acmlicensed}
\acmConference[CIKM '22]{Proceedings of the 31st ACM International Conference on Information and Knowledge Management}{October 17--21, 2022}{Atlanta, GA, USA}
\acmBooktitle{Proceedings of the 31st ACM International Conference on Information and Knowledge Management (CIKM '22), October 17--21, 2022, Atlanta, GA, USA}
\acmPrice{15.00}
\acmDOI{10.1145/3511808.3557399}
\acmISBN{978-1-4503-9236-5/22/10}
\begin{document}

% \title[Joint time and mark]{Modeling Inter-Dependence Between Time and Mark in Multivariate Temporal Point Processes}
\title[Modeling Inter-Dependence Between Time and Mark in Multivariate Temporal Point Processes]{Modeling Inter-Dependence Between Time and Mark \\ in Multivariate Temporal Point Processes}

% \affiliation{%

% \affiliation{%

\author{Govind Waghmare}
\affiliation{%
  \institution{Mastercard, AI Garage}
  \city{Gurugram}
  \country{India}}
\email{govind.waghmare@mastercard.com}

\author{Ankur Debnath}
\affiliation{%
  \institution{Mastercard, AI Garage}
   \city{Gurugram}
  \country{India}}
\email{ankur.debnath@mastercard.com}

\author{Siddhartha Asthana}
\affiliation{%
  \institution{Mastercard, AI Garage}
   \city{Gurugram}
  \country{India}}
\email{siddhartha.asthana@mastercard.com}

\author{Aakarsh Malhotra}
\affiliation{%
  \institution{Mastercard, AI Garage}
   \city{Gurugram}
  \country{India}}
\email{aakarsh.malhotra@mastercard.com}

% \affiliation{%

% \affiliation{%

% \affiliation{%

% \affiliation{%

% \affiliation{%

% \renewcommand{\shortauthors}{Waghmare, et al.}
\renewcommand{\shortauthors}{Govind Waghmare, Ankur Debnath, Siddhartha Asthana, \& Aakarsh Malhotra}
\begin{abstract}
  Temporal Point Processes (TPP) are probabilistic generative frameworks. They model discrete event sequences localized in continuous time. Generally, real-life events reveal descriptive information, known as marks. Marked TPPs model time and marks of the event together for practical relevance. Conditioned on past events, marked TPPs aim to learn the joint distribution of the time and the mark of the next event. For simplicity, conditionally independent TPP models assume time and marks are independent given event history. They factorize the conditional joint distribution of time and mark into the product of individual conditional distributions. This structural limitation in the design of TPP models hurt the predictive performance on entangled time and mark interactions. In this work, we model the conditional inter-dependence of time and mark to overcome the limitations of conditionally independent models. We construct a multivariate TPP conditioning the time distribution on the current event mark in addition to past events. Besides the conventional intensity-based models for conditional joint distribution, we also draw on flexible intensity-free TPP models from the literature. The proposed TPP models outperform conditionally independent and dependent models in standard prediction tasks. Our experimentation on various datasets with multiple evaluation metrics highlights the merit of the proposed approach.
\end{abstract}





\maketitle

\begin{figure}[ht]
    \begin{center}
        \includegraphics[width=\linewidth]{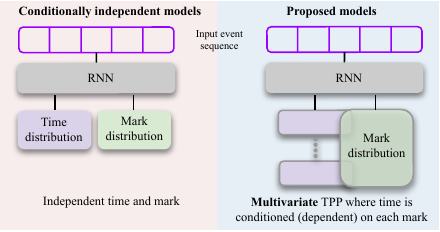}
        % \caption{Marked TPP model classification at a broad level based on the dependence relationship between time and mark in the conditional joint distribution}
        \caption{The proposed models are conditionally dependent, multivariate, and capable of employing both intensity-free and intensity-based formulations.}
        \label{fig:visual_abstract}
  \end{center}
\end{figure}

\section{Introduction}

TPP is a random process representing irregular event sequences occurring in continuous time. Financial transactions, earthquakes, and electronic health records (EHR) exhibit asynchronous temporal patterns. TPPs are well studied in the literature and have rich theoretical foundations \cite{alex_hawkes, cramer, brockmeyer1948life}.  Classical (non-neural) TPPs focus on capturing relatively simple temporal patterns through Poison process \cite{kingman1992poisson}, self-excitation process \cite{alex_hawkes}, and self-correcting process \cite{ISHAM1979335}. With the advent of neural networks, many flexible and efficient neural architectures have been developed to model multi-modal event dynamics, called neural TPPs \cite{shchur2021neural}.

Any attribute associated with an event makes it more realistic and represented as a mark. Marks capture a better description of the event, like time and location, interacting entities, and their evolution. Stochastic modeling of such events to study underlying event generation mechanisms is called the marked TPPs. For instance, in seismology, earthquake event dynamics are better understood with the knowledge of magnitude and location \cite{chen2021neuralstpp}. A temporal model solely learned on time may not be of practical relevance where marks impart realistic and reliable information. Marked TPP is a probabilistic framework \cite{daley2007introduction} which aims to model the joint distribution of time and mark of the next event using previous event history. An estimation of the next event time and the mark has practical application in many domains that exhibit complex time and mark interactions. Such application include online user engagements \cite{farajtabar2014shaping, sharma_amdn_hage_kdd_21, zhang_vigdet_21}, information diffusion \cite{uncovering_rodriguez}, econometrics \cite{bacry2015market}, and healthcare \cite{enguehard2020neural}. In personalized healthcare, a patient could have a complex medical history, and several diseases may depend on each other. Predictive EHR modeling could reveal potential future clinical events and facilitate efficient resource allocation.

\noindent \textbf{Time and mark dependency}: While modeling the conditional joint distribution of time and marks, many prior works assume marks to be conditionally independent of time \cite{du2016recurrent,omi_FNN}. This assumption on the conditional joint distribution of time and mark leads to two types of marked TPPs, (i) conditionally independent, and (ii) conditionally dependent models. The independence assumption allows factorization of the conditional joint distribution into a product of two independent conditional distributions. It is the product of continuous-time distribution and categorical mark distribution\footnote{Categorical marks are conventional in the prior works.}, both conditioned on the event history. The independence between time and mark limits the structural design of the neural architecture in conditionally independent models. Thus, such models require fewer parameters to specify the conditional joint distribution of time and marks but fail to capture their dependence. On the contrary, conditionally dependent models capture the dependency between time and mark by either conditioning time distribution on mark or mark distribution on time. A recent study by \cite{enguehard2020neural} shows that the conditionally independent models perform poorly compared to conditionally dependent models.

\noindent \textbf{Multivariate TPP}: Marked TPP is a joint probability distribution over a given time interval. In order to model time and mark dependency, the time distribution should be conditioned on all possible marks. It leads to a multivariate TPP model where a tuple of time distributions is learned over a set of categorical marks \cite{mei2017neuralhawkes}. For $K$ distinct marks, $k^{th}$ multivariate distribution ($k \in \{1, \dots, K \}$) indicates the joint distribution of the time and the $k^{th}$ mark. 

\noindent \textbf{Intensity-based vs intensity-free modeling}: In both conditionally independent and conditionally dependent models, inter-event time distribution is a key factor of the joint distribution. The standard way of learning time distribution is by estimating conditional intensity function. However, the intensity function requires selecting good parametric formulation \cite{shchur2020intensity}. The parametric intensity function often makes assumptions about the latent dynamics of the point process. A simple parametrization has limited expressiveness but makes likelihood computation easy. Though an advanced parametrization adequately captures event dynamics, likelihood computation often involves numerical approximation using Newton-Raphson or Monte Carlo (MC). Besides intensity-based formulation, other ways to model conditional inter-event time distribution involve probability density function (PDF) modeling, cumulative distribution function, survival function, and cumulative intensity function \cite{shchur2021neural, Okawa_2019}. A model based on an intensity-free focuses on closed-form likelihood, closed-form sampling, and flexibility to approximate any distribution.

In this work, we model inter-dependence between time and mark by learning conditionally dependent distribution. While inferring the next event, we model a PDF of inter-event time distribution for each discrete mark. The time distribution conditioned on marks improves the predictive performance of the proposed models compared to others. A high-level overview of our approach is shown in Figure \ref{fig:visual_abstract}. In summary, we make the following contributions:

\begin{itemize}
    \item We overcome the structural design limitation of conditionally independent models by proposing novel \textit{conditionally dependent, both intensity-free and intensity-based,} and \textit{multivariate} TPP models. To capture inter-dependence between mark and time, we condition the time distribution on the current mark in addition to event history. 
    \item We improve the predictive performance of the intensity-based models through conditionally dependent modeling. Further, we draw on the intensity-free literature to design a flexible multivariate marked TPP model. We model the PDF of conditional inter-event time to enable closed-form likelihood computation and closed-form sampling. 
    \item Using multiple metrics, we provide a comprehensive evaluation of a diverse set of synthetic and real-world datasets. The proposed models consistently outperform both conditionally independent and conditionally dependent models. 
\end{itemize}

\section{Related work}
In this section, we provide a brief overview of classical (non-neural) TPPs and neural TPPs. Later, we discuss conditionally independent and conditionally dependent models. In the end, we differentiate the proposed solution against state-of-the-art models in the literature.

\subsection{Classical (non-neural) TPPs} TPPs are mainly described via conditional intensity function. Basic TPP models make suitable assumptions about the underlying stochastic process resulting in constrained intensity parametrizations. For instance, Poisson process \cite{kingman1992poisson, palm1943intensiatsschwankungen} assumes that inter-event times are independent. In Hawkes process \cite{alex_hawkes_cluster, Ogata1998} event excitation is positive, additive over time, and decays exponentially with time. Self-correcting process \cite{ISHAM1979335} and autoregressive conditional duration process \cite{Engle1998AutoregressiveCD} propose different conditional intensity parametrizations to capture inter-event time dynamics. These constraints on conditional intensity limit the expressive power of the models and hurt predictive performance due to model misspecification \cite{du2016recurrent}. 

\subsection{Neural TPPs}
Neural TPPs are more expressive and computationally efficient than classical TPPs due to their ability to learn complex dependencies. A TPP model inferring the time and mark of the next event \textit{sequentially} is called autoregressive (AR) TPP. A seminal work by \cite{du2016recurrent, xiao2017modeling} connects the point processes with a neural network by realizing conditional intensity function using a recurrent neural network (RNN). Generally, the event history is encoded using either recurrent encoders or set aggregation encoders \cite{zhang2020self, zuo2020transformer}.

\noindent{\textbf{Conditionally independent models}} assume time and mark are independent and inferred from the history vector representing past events. This assumption makes this neural architecture computationally inexpensive but hurts the predictive performance as the influence of mark and time on each other cannot be modeled. Therefore, all conditional independent models perform similarly on mark prediction due to their limited expressiveness \cite{shchur2020intensity}. Further, modeling time distribution based on conditional intensity is conventional in multiple prior models \cite{du2016recurrent, xiao2017modeling}. The training objective in these models involves numerical approximations like MC estimates. On the contrary, \cite{shchur2020intensity} proposed intensity-free learning of TPPs where PDF of inter-event times is learned directly (bypassing intensity parametrization) via log-normal mixture (LNM) distribution. LNM model focuses on flexibility, closed-form likelihood, and closed-form sampling. 

\begin{table}[t]
  \centering
    \caption{Comparison of the proposed models with other neural temporal point processes. }
    \begin{tabular}{cccc}
    \toprule
    Model & \begin{tabular}{@{}c@{}}Conditionally \\ dependent \\modeling\\ \end{tabular} & \begin{tabular}{@{}c@{}}Intensity-free \\ modeling \\\end{tabular} \\
    \midrule
    
    Conditional Poisson (CP) & {\color{red}\XSolidBrush} & {\color{red}\XSolidBrush}\\
    RMTPP (\cite{du2016recurrent}) & {\color{red}\XSolidBrush} & {\color{red}\XSolidBrush}\\
    LNM (\cite{shchur2020intensity}) & {\color{red}\XSolidBrush} & {\color{green}\checkmark}\\
    NHP (\cite{mei2017neuralhawkes}) & {\color{green}\checkmark} & {\color{red}\XSolidBrush}\\
    SAHP (\cite{zhang2020self}) & {\color{red}\XSolidBrush} & {\color{red}\XSolidBrush}\\
    THP (\cite{zuo2020transformer}) & {\color{red}\XSolidBrush} & {\color{red}\XSolidBrush}\\
    \midrule
    \textit{\textbf{Proposed RMTPP}} & {\color{green}\checkmark} & {\color{red}\XSolidBrush}\\
    \textit{\textbf{Proposed LNM}} & {\color{green}\checkmark} & {\color{green}\checkmark} \\
    \textit{\textbf{Proposed THP}} & {\color{green}\checkmark} & {\color{red}\XSolidBrush} \\
    
    \bottomrule
  \end{tabular}
  \label{tab:literature_comparison}
\end{table}

\noindent{\textbf{Conditionally dependent models}} capture dependency either by conditioning time on marks \cite{zuo2020transformer, enguehard2020neural, mei2017neuralhawkes} or marks on time \cite{bilos_nips_2019}. In \cite{enguehard2020neural, mei2017neuralhawkes}, a separate intensity function is learned for each mark at every time step making it multivariate TPP. \cite{mei-2020-nce, Trkmen2019FastPointSD, Guo2018INITIATORNE} discuss the scalability of models when the number of marks is large. These models are intensity-based and hence share the same drawbacks discussed previously compared to intensity-free.

In the proposed models, we realize the conditional dependence of time and marks. We rely on both standard intensity-based and intensity-free formulations to realize PDF of inter-event time. For intensity-based case, we draw on well-known models like RMTPP \cite{du2016recurrent} and THP \cite{zuo2020transformer}, called as \textit{\textbf{proposed RMTPP}} and \textit{\textbf{proposed THP}} respectively. The intensity-free model allows analytical (closed-form) computation of likelihood. We draw on the conditionally independent log-normal mixture (LNM) model proposed by \cite{shchur2020intensity} to design a conditionally dependent multivariate TPP model known as \textit{\textbf{proposed LNM}}.  The advantage of the proposed methods compared to the state-of-the-art models is shown in Table \ref{tab:literature_comparison}. 

\section{Model Formulation}
\subsection{Background and notations}

We represent variable-length event sequence with time and mark attributes as $E = \{ e_1=(t_1, m_1), \dots, e_N=(t_N, m_N) \}$ over time interval $[0,T]$ where $ t_1 < \dots < t_N $ are event arrival times and $ m_i \in \mathcal{M} $ are categorical marks from the set $\mathcal{M} = \{ 1, 2, \dots, K \} $. The number of events $N$ in the interval $[0,T]$ is a random variable. The inter-event times are given as $ \tau_i = t_i - t_{i-1}$ with $t_0 = 0$ and $t_{N+1} = T$. The event history till the time $t$ is stated as $ \mathcal{H}_{t} = \{(t_i,m_i) : t_i < t \} $. The joint distribution of the next event conditioned on past events is defined as $P(t_i, m_i \vert \mathcal{H}_{t_i}) = P^{*}(t_i, m_i)$. Here, $*$ symbol here indicates the joint distribution is conditioned on the event history $\mathcal{H}_{t_i}$ \cite{daley2007introduction}. Ordinarily, multivariate TPP with K categorical marks is characterized by conditional intensity function $\lambda^{*}_{k}(t)$ for the event of type $k$. It is defined as 
\begin{equation}\label{eqn:marked_intensity}
\lambda^{*}_{k}(t)=  \lim_{dt\to 0} \frac{\text{Pr(event of type } k \text{ in } [t, t+ dt) \vert \mathcal{H}_t)}{dt}
\end{equation}
For unmarked case, number of marks, K=1 and Equation \ref{eqn:marked_intensity} becomes $\lambda^{*}_{k}(t)= \lambda^{*}(t)$. Here,  $\lambda^{*}(t)$ is called as \textit{ground intensity} \cite{rasmussen2011temporal}. The conditional inter-event time PDF for the  $i^{th}$ event of type $k$ is given as
\begin{equation}\label{eqn:time_PDF_for_mark}
  f^{*}_{ik}(\tau_i) = \lambda^{*}_{k}(t_{i-1} + \tau_{i}) \exp \left(-\sum_{k=1}^{K} \int_{t_{i-1}}^{t_{i}} \lambda_{k}^{*}(t^{'})dt^{'} \right)
\end{equation}
Here, $\tau_{i} = t_{i}-t_{i-1}$ indicates inter-event time is isomorphic with arrival-time and could be used interchangeably.

\noindent \textbf{Conditionally independent models} factorize the conditional joint distribution $P^{*}_{i}(\tau_{i}, m_i)$ into product of two independent distributions  $P^{*}_{i}(\tau_{i})$ and $ P^{*}_{i}(m_{i})$. The conditional joint density\footnote{We use the conditional density term in the broad sense. Here, time is continuous random variable and mark is discrete random variable \cite{rasmussen2011temporal}.} of the time and the mark is represented as 
\begin{equation}\label{eqn:joint_dist_cond_ind_models}
f^{*}_{i}(\tau_{i}, m_{i})=f^{*}_{i}(\tau_{i}) \cdot p^{*}_{i}(m_{i}),
\end{equation}
where, $f^{*}_{i}(\tau_{i})$ is PDF of the time distribution $P^{*}_{i}(\tau_{i})$ and $p^{*}_{i}(m_{i})$ is the probability mass function (PMF) of categorical mark distribution $P^{*}_{i}(m_{i})$. Now, there are two ways to model the time PDF $f^{*}_{i}(\tau_{i})$. One way is to use conditional intensity function as follows
\begin{equation}\label{eqn:time_PDF_ground_intensity}
  f^{*}_{i}(\tau_{i}) = \lambda^{*}(t_{i-1} + \tau_{i}) \exp \left(-\int_{t_{i-1}}^{t_{i}} \lambda^{*}(t^{'})dt^{'} \right),
\end{equation}
and other is parametric density estimation of PDF $f^{*}_{i}(\tau_{i})$ using history $\mathcal{H}_{t_i}$. In conditionally independent models, time distribution is not conditioned on the current mark. 
So, $f^{*}_{i}(\tau_{i} \vert m_{i}) = f^{*}_{i}(\tau_{i})$ and the model does not capture the influence of the current mark on time distribution.

\noindent \textbf{Conditionally dependent models} capture the dependency between $\tau_{i}$ and $m_i$ either by conditioning time on mark or by conditioning mark on time. When time is conditioned on marks, a separate distribution $P^{*}_{i}(\tau_{i} \vert m_{i} = k)$ is specified for each mark $k \in \mathcal{M} $. Here, the conditional joint density for each mark takes the following form:
\begin{equation}\label{eqn:time_condtioned_on_marks}
f^{*}_{i}(\tau_{i}, m_{i}=k)=f^{*}_{i}(\tau_{i} \vert m_{i}=k) \cdot p^{*}_{i}(m_{i}=k),
\end{equation}
Usually, the time PDF $f^{*}_{i}(\tau_{i} \vert m_{i}=k)=f^{*}_{ik}(\tau_i) $ is represented using parametrized intensity function (Equation \ref{eqn:time_PDF_for_mark}). When marks are conditioned on the time, a distribution $P^{*}_{i}(m_{i} \vert \tau_{i} = \tau)$ is specified for all values of $\tau$. \cite{bilos_nips_2019} parametrized the distribution $P^{*}_{i}(m_{i} \vert \tau_{i} = \tau)$ using Gaussian process. Here, the joint density at $k^{th}$ mark is given as 
\begin{equation}
f^{*}_{i}(\tau_{i}, m_{i}=k)=f^{*}_{i}(\tau_{i}) \cdot p^{*}_{i}(m_{i} = k \vert \tau_i=t-t_{i-1}),
\end{equation}
where, $t_{i-1} < t \leq t_{i}$ and time PDF $f^{*}_{i}(\tau_{i})$ is generally obtained using Equation \ref{eqn:time_PDF_ground_intensity}.

\begin{figure*}[t]
  \centering
  \includegraphics[width=\linewidth]{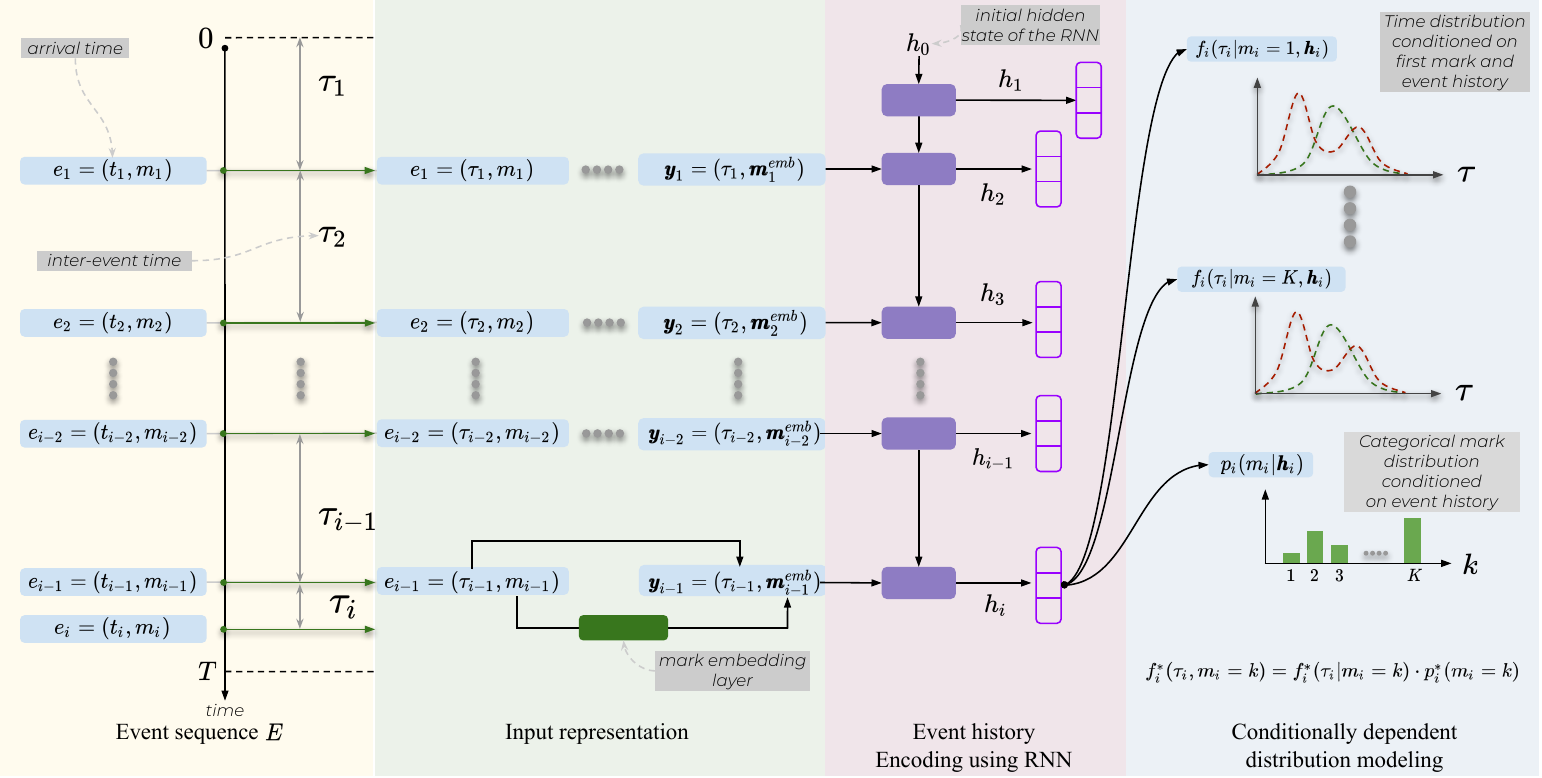}
  \caption{Overview of the proposed multivariate conditionally dependent model. The inter-event time distribution is learned either using an intensity-free or intensity-based approach. Input event sequence contains arrival time and mark for each event. Input representation contains inter-event time and mark embedding. RNN converts event history into a fixed dimension vector. In the end, we compute the conditional joint density of time and marks.}
  \label{fig:architecture}
\end{figure*}

\subsection{Proposed approach}
We model the conditional joint distribution of the time and the mark by conditioning time on marks (Equation \ref{eqn:time_condtioned_on_marks}). We specify inter-event time PDF conditioned on each mark type. A schematic representation of the proposed approach is shown in Figure \ref{fig:architecture}. Note that, the proposed approach is common for both intensity-based and intensity-free models. For intensity-based models, time PDF $f^{*}_{ik}(\tau_i)$ in Equation \ref{eqn:time_condtioned_on_marks} is realized using Equation \ref{eqn:time_PDF_for_mark}. For both \textit{proposed RMTPP} and \textit{proposed THP} models, we use parametrized intensity functions defined in the papers \cite{du2016recurrent} and \cite{zuo2020transformer} respectively. We condition these intensity functions on marks (Equation \ref{eqn:time_PDF_for_mark}) to alleviate the structurally independent time and mark assumption. Intensity-free based approach (\textit{proposed LNM}) is explained further. A parametric density estimation approach based on the log-normal mixture (LNM) is proposed by \cite{shchur2020intensity} for conditionally independent models. We draw on this approach to design multivariate TPP capable of capturing inter-dependence between time and marks. We realize a conditional joint density $f^{*}_{i}(\tau_{i}, m_{i}=k)$ of the $i^{\text{th}}$ event with $k^{th}$ type using event history $\mathcal{H}_{t_{i}}$ till $(i-1)^{\text{th}}$ event. For a given variable length event sequence $E$, each event is represented as $e_j=(t_{j},m_{j})$. Categorical marks are encoded using embedding function as $\pmb{m}^{emb}_{j}=\textit{Embedding}(m_{j})$. Here, the embedding function is a learnable matrix $ \pmb{E} \in \mathbb{R}^{K \times \vert \pmb{m}^{emb}_{j} \vert}$ and $\pmb{m}^{emb}_{j}=\text{one-hot}(m_j) \cdot \pmb{E}$.
We concatenate inter-event time $\tau_j$ and mark embedding $\pmb{m}^{emb}_{j}$ to form input feature $\pmb{y_{j}} = (\tau_{j}, \pmb{m}^{emb}_{j})$. The RNN converts the input representation $(\pmb{y}_{1}, \pmb{y}_{2}, \dots, \pmb{y}_{i-1})$ into fixed-dimensional history vector $\pmb{h}_{i}$. Here, $\mathcal{H}_{t_i} = \pmb{h}_{i}$. Starting with initial hidden state $\pmb{h}_0$, next hidden state of the RNN is updated as $\pmb{h}_{i}=\textit{Update}(\pmb{h}_{i-1}, \pmb{y}_{i-1})$. For conditionally dependent multivariate TPP, we learn PDF of inter-event time using log-normal mixture model as follows:
\begin{equation}
    f^{*}(\tau \vert m=k)=f^{*}_{k}(\tau)=f_{k}(\tau \vert \pmb{w}_{k}, \pmb{\mu}_{k}, \pmb{s}_{k}),
\end{equation}

where,  $\pmb{w}$ are the mixture weights, $\pmb{\mu}$ are the mixture means and $\pmb{s}$ are the mixture standard deviations. Further, inline with \cite{shchur2020intensity},
% $f_{k}(\tau \vert \pmb{w}_{k}, \pmb{\mu}_{k}, \pmb{s}_{k})$ is formulated as 
\begin{equation}
    f_{k}(\tau \vert \pmb{w}_{k}, \pmb{\mu}_{k}, \pmb{s}_{k}) = \sum_{c=1}^{C} w_{k,c} \frac{1}{\tau s_{k,c} \sqrt{2 \pi}} \exp \left(  -\frac{(\log {\tau} - \mu_{k,c})^{2}}{2s_{k,c}^2} \right) , 
\end{equation}
where, $c \in \{1,2, \dots, C \} $ indicates number of mixture components. We discuss the selection of $C$ and its impact on the result in Table \ref{tab:mixture_comp_sensitivity}. For each mark $k \in \mathcal{M}$, the parameters $\pmb{w}_{k}, \pmb{\mu}_{k} \text{ and } \pmb{s}_{k}$ are estimated from $\pmb{h}$ as follows\footnote{subscript $i$ (event index) is dropped for simplicity}: 
\begin{equation}
    \pmb{w}_{k} = \text{softmax} (\pmb{W}_{\pmb{w}_{k}} \pmb{h} + \pmb{b}_{\pmb{w}_{k}})
\end{equation}
\begin{equation}
    \pmb{\mu}_{k} = \exp (\pmb{W}_{\pmb{\mu}_{k}} \pmb{h} + \pmb{b}_{\pmb{\mu}_{k}})
    \text{ and }
    \pmb{s}_{k} = \pmb{W}_{\pmb{s}_{k}} \pmb{h} + \pmb{b}_{\pmb{s}_{k}}
\end{equation}    
% \begin{equation}
%     \pmb{s}_{k} = \pmb{W}_{\pmb{s}_{k}} \pmb{h} + \pmb{b}_{\pmb{s}_{k}}
% \end{equation}
% $\pmb{w}_{k} = \text{softmax} (\pmb{W}^{\pmb{w}_{k}} \pmb{h}_{i} + \pmb{b}^{\pmb{w}_{k}})$, $\pmb{\mu}_{k} = \exp (\pmb{W}^{\pmb{\mu}_{k}} \pmb{h}_{i} + \pmb{b}^{\pmb{\mu}_{k}})$ and $\pmb{s}_{k} = \pmb{W}^{\pmb{s}_{k}} \pmb{h}_{i} + \pmb{b}^{\pmb{s}_{k}}$.
Here, $ \{ \pmb{W}_{\pmb{w}_{k}}, \pmb{W}_{\pmb{\mu}_{k}}, \pmb{W}_{\pmb{s}_{k}}, \pmb{b}_{\pmb{w}_{k}}, \pmb{b}_{\pmb{\mu}_{k}}, \pmb{b}_{\pmb{s}_{k}} \} $ are the learnable parameters of the neural network. We parametrize the categorical mark distribution for mark prediction. The history vector $\pmb{h}$ is passed through linear layer with weight matrix $\pmb{W}_{m} = [\pmb{w}_1, \dots, \pmb{w}_K] $ and bias vector $\pmb{b}_{m} = [b_1, \dots, b_K]$. Here, $\pmb{W}_{m} \in \mathbb{R}^{\vert \pmb{h} \vert \times K}$. The mark distribution is computed using softmax function as follows:

\begin{equation}
    p(m=k \vert \pmb{h}) = p^{*}(m=k) = \frac{\exp (\pmb{w}^{\top}_k \pmb{h} + b_{k})}{\sum_{j=1}^{K} \exp (\pmb{w}^{\top}_j \pmb{h} + b_{j})}
\end{equation}

\begin{table*}[t]
    \centering
    
    \caption{Dataset statistics and hyperparameters}
    
    \begin{tabular}{c|ccccc|cccccccc}
    
    \toprule
      
    & \multicolumn{5}{c|}{\textbf{Statistics}} 
    & \multicolumn{8}{c}{\textbf{Hyperparameters}} \\ 
        
    \cmidrule(r){2-6} 
    \cmidrule(r){7-14} 
        
    & \begin{tabular}{@{}c@{}}\#Marks\\ \end{tabular} 
    & \begin{tabular}{@{}c@{}}\#Seq.\end{tabular} 
    & \begin{tabular}{@{}c@{}}\#Events\end{tabular} 
    & \begin{tabular}{@{}c@{}}Start \\ time\end{tabular}
    & \begin{tabular}{@{}c@{}}End \\ time\end{tabular}
    & \begin{tabular}{@{}c@{}}Train \\ size \end{tabular} 
    & \begin{tabular}{@{}c@{}}Val \\ size \end{tabular}
    & \begin{tabular}{@{}c@{}}Test \\ size \end{tabular}
    & \begin{tabular}{@{}c@{}}Batch \\ size \end{tabular}
    & \begin{tabular}{@{}c@{}}Mixture \\ comp \\C \end{tabular}
    & \begin{tabular}{@{}c@{}}History \\ vector \\ size \end{tabular}
    & \begin{tabular}{@{}c@{}}Mark \\ emb \\ size \end{tabular}
    & \begin{tabular}{@{}c@{}}Time \\ scale \end{tabular}
    \\
     
    \midrule
    
    \begin{tabular}{@{}c@{}}Hawkes\\Ind. \end{tabular} 	& 2 & 24576 & 457788 & 0 & 100 & 14745 & 4915 & 4916 & 512 & 64 & 64 & 32 & 1 \\
    
    \hline
    
    \begin{tabular}{@{}c@{}}Hawkes\\Dep. (I) \end{tabular} 	& 2 & 24576 & 607512 & 0 & 100 & 14745 & 4915 & 4916 & 512 & 64 & 64 & 32 & 1 \\
    
    \hline
    
    \begin{tabular}{@{}c@{}}Hawkes\\Dep. (II) \end{tabular} 	& 5 & 30000 & 12741668 & 0 & 100 & 18000 & 6000 & 6000 & 512 & 64 & 64 & 32 & 1 \\
    
    \hline
    
    \begin{tabular}{@{}c@{}}MIMIC\\ II \end{tabular} 	& 75 & 715 & 2419 & 0 & 6 & 429 & 143 & 143 & 64 & 64 & 64 & 32 & 1 \\
    
    \hline
    
    MOOC 	& 97 & 7047 & 396633 & 0 & $25.73e5$ & 4228 & 1409 & 1410 & 64 & 64 & 64 & 64 & 1 \\
    
    \hline
    
    \begin{tabular}{@{}c@{}}Stack\\Overflow  \end{tabular} 	& 22 & 6633 & 480413 & $1.32e9$ & $1.38e9$ & 3979 & 1327 & 1327 & 64 & 64 & 64 & 32 & $1.e-05$ \\
    
    \bottomrule
    
    \end{tabular}
  
  \label{tab:parameters_of_datasets}
\end{table*}

\noindent In conditionally independent models, $f^{*}_{i}(\tau_{i}, m_{i}=k) = f^{*}_{i}(\tau_{i}) \cdot p^{*}_{i}(m_{i}=k)$. As $f^{*}_{i}(\tau_{i})$ is independent of mark, estimation of mark is done as follows:
\begin{equation}\label{eqn:mark_cond_independent}
    \argmax_{k \in \mathcal{M}} f^{*}_{i}(\tau_{i}, m_{i}) \coloneqq \argmax_{k \in \mathcal{M}} p^{*}_{i}(m_{i}=k)
\end{equation}

\noindent On the other hand, in conditionally dependent models, mark is estimated as follows:
\begin{equation}\label{eqn:mark_cond_dependent}
    \argmax_{k \in \mathcal{M}} f^{*}_{i}(\tau_{i}, m_{i}) \coloneqq \argmax_{k \in \mathcal{M}} f^{*}_{i}(\tau \vert m=k) \cdot p^{*}_{i}(m_{i}=k)
\end{equation}

\subsection{Likelihood estimation}

As neural TPP is a generative framework, maximum likelihood estimation (MLE) is a widely used training objective. Other objectives could be inverse reinforcement learning \cite{upa_RL_2018, li2018learning}, Wasserstein distance \cite{xiao_wasserstein_2017, deshpande2021long} and adversarial losses \cite{wu_2018, yan_ijcai2018}. In the proposed approaches we use the MLE objective. For the event sequence $E = \{ e_1=(t_1, m_1), \dots, e_N=(t_N, m_N) \}$ in the interval $[0,T]$, the likelihood function represents the joint density of all the events. So, likelihood is factorized into the product of conditional joint densities (of time and mark) for each event. The negative log-likelihood (NLL) is formulated as:

\begin{multline}\label{eqn:NLL_computation}
    - \log p(E) = -\sum_{i=1}^{N} \sum_{k=1}^{K} \mathbbm{1}_{(m_{i}=k)} \log f_{i}(\tau_{i}, m_{i} = k \vert \mathcal{H}_{t_{i}}) \\ - \log(1-P(T \vert \mathcal{H}_{T})), 
\end{multline}

where, $f_{i}(\tau_{i}, m_{i} = k \vert \mathcal{H}_{t_i})$ is the conditional joint density of event with mark type $k$ and $(1-P(T \vert \mathcal{H}_{T})$ indicates no event of any type has occurred in the interval between $t_N$ and $T$ (survival probability of the last interval). As \textit{proposed RMTPP} and \textit{proposed THP} use conditional intensity-based NLL formulation,  they requires approximation of the integral in the Equation \ref{eqn:marked_intensity} using MC \cite{mei2017neuralhawkes, du2016recurrent}. In the \textit{proposed LNM}, mixture model enables computation of NLL analytically which is more accurate and computationally efficient than MC approximations \cite{shchur2021neural}. For NLL computation, we factorize $f_i(\tau_{i}, m_{i} = k \vert \mathcal{H}_{t_i})$ according to the Equation \ref{eqn:time_condtioned_on_marks}. 

\section{Experimental Evaluation}

\subsection{Datasets}
We perform experiments on commonly used synthetic and real-world benchmark datasets in the marked TPP literature. All datasets contain multiple unique sequences and show variations in the sequence length. We include three synthetic datasets and three real-world datasets for experimentation. Dataset details and summary statistics are given in the Table \ref{tab:parameters_of_datasets}.

\subsubsection{Synthetic datasets}
Using Hawkes dependent and independent processes, we generate three datasets. These datasets are commonly used in state-of-the-art models like \cite{shchur2020intensity, omi_FNN, enguehard2020neural}. Hawkes process is self-exciting point process with following conditional intensity function represented as \footnote{\href{https://x-datainitiative.github.io/tick/modules/generated/tick.hawkes.SimuHawkesExpKernels.html\#tick.hawkes.SimuHawkesExpKernels}{We use tick library to generate Hawkes datasets}}:
\begin{equation}
    \lambda_{k}^*(t) = \mu_k +  \sum_{j=1}^{K} \sum_{i:t_{j,i} < t} \alpha_{k,j} \beta_{k,j} \exp(-\beta_{k,j}(t-t_{j,i}))
\end{equation}

Here, $\mu_k$ is base intensity, $\alpha_{k,j}$ is excitation (intensity) between event types and $\beta_{k,j}$ is a decay of the exponential kernel. Inline with \cite{enguehard2020neural, mei2017neuralhawkes},  using different values of $\pmb{\mu}, \pmb{\alpha}$ and $\pmb{\beta}$, we generate Hawkes independent (denoted as Hawkes Ind.) and Hawkes dependent dataset (denotes as Hawkes Dep. (I)). Hawkes Ind. and Hawkes Dep. (I) are comparatively simple datasets. Therefore, we also generate another Hawkes dependent dataset (denoted as Hawkes Dep. (II) with five different marks and longer average sequence length to make prediction challenging (see Table \ref{tab:parameters_of_datasets}). For the Hawkes Ind. dataset, we use the following parameters:

\begin{equation} 
            \pmb{u} = \begin{bmatrix}
                    0.1 & 0.05 
            \end{bmatrix} 
            \quad
            \pmb{\alpha} = \begin{bmatrix}
                0.2 & 0.0 \\
                0.0 & 0.4
                \end{bmatrix}
                \quad
                \pmb{\beta} = \begin{bmatrix}
                1.0 & 1.0 \\
                1.0 & 2.0
                \end{bmatrix}
\end{equation}

For Hawkes Dep. (I) dataset, we use following parameters:

\begin{equation} 
    \pmb{u} = \begin{bmatrix}
                    0.1 & 0.05 
            \end{bmatrix} 
            \quad
             \pmb{\alpha} = \begin{bmatrix}
                0.2 & 0.1 \\
                0.2 & 0.3
                \end{bmatrix}
                \quad
                \pmb{\beta} = \begin{bmatrix}
                1.0 & 1.0 \\
                1.0 & 1.0
                \end{bmatrix}
\end{equation}

For Hawkes Dep. (II) dataset, we randomly sample parameters inline with \cite{mei2017neuralhawkes} as follows:

\begin{equation} 
    \pmb{u} = \begin{bmatrix}
                    0.713 & 0.057 & 0.844 & 0.254 & 0.344 
            \end{bmatrix} 
\end{equation}

\begin{equation}   
    \pmb{\alpha} = \begin{bmatrix}
                0.689 & 0.549 & 0.066 & 0.819 & 0.007 \\
                0.630 & 0.000 & 0.457 & 0.622 & 0.141 \\
                0.134 & 0.579 & 0.821 & 0.527 & 0.795 \\
                0.199 & 0.556 & 0.147 & 0.030 & 0.649 \\
                0.353 & 0.557 & 0.892 & 0.638 & 0.836 
                \end{bmatrix}
\end{equation}
                
\begin{equation} \pmb{\beta} = \begin{bmatrix}
                9.325 & 9.764 & 2.581 & 4.007 & 9.319 \\
                5.759 & 8.742 & 4.741 & 7.320 & 9.768 \\
                2.841 & 4.349 & 6.920 & 5.640 & 3.839 \\
                6.710 & 7.460 & 3.685 & 4.052 & 6.813 \\
                2.486 & 2.214 & 8.718 & 4.594 & 2.642 
  
                \end{bmatrix}
\end{equation}

\subsubsection{Real-world datasets}
For real-world datasets, we use publicly available common benchmark datasets like Stack Overflow\footnote{https://archive.org/details/stackexchange} \citep{du2016recurrent}, MOOC\footnote{https://github.com/srijankr/jodie/} \citep{kumar2019predicting} and MIMIC-II\footnote{https://github.com/babylonhealth/neuralTPPs} \citep{enguehard2020neural}.
Stack Overflow is a question-answering website. Users on this site earn badges as a reward for contribution. For each user, the event sequence represents different badges received over two years. MOOC dataset captures interactions of learners with the online course system. Different actions like taking a course and solving an assignment are different kinds of marks. MIMIC-II dataset contains anonymized electronic health records of the patients visiting the intensive care unit for seven years. Each event represents the time of the hospital visit. The mark indicates the type of disease (75 unique diseases). Further dataset statistics including the number of events, event start time, and event end time are shown in Table \ref{tab:parameters_of_datasets}.

\begin{table*}[t]
    \centering
    
    \caption{Predictive performance of marked TPP models. NLL/time is normalized NLL score over event sequence interval. For marks, we report micro F1 score and weighted F1 score (denoted as Wt. F1 score). Bold numbers indicate the best performance. Results on the remaining datasets are provided in the Tables \ref{tab:results_table_2} and \ref{tab:results_table_3}. Prop. stands for Proposed.}
    
     \begin{tabular}{c|cccccc|cccccc}
  
    \toprule
      
        & \multicolumn{6}{c|}{\textbf{Hawkes Dependent I (Synthetic dataset)}} & \multicolumn{6}{c}{\textbf{MOOC (Real dataset)}} \\ 
        \cmidrule(r){2-7} 
        \cmidrule(r){8-13} 
        Model 
        & \begin{tabular}{@{}c@{}}Time\\NLL \end{tabular}
        & \begin{tabular}{@{}c@{}}Mark\\NLL \end{tabular}
        & \begin{tabular}{@{}c@{}}NLL \end{tabular} 
        & \begin{tabular}{@{}c@{}}NLL/ \\ Time \end{tabular}  
        & \begin{tabular}{@{}c@{}}Micro\\F1 \end{tabular}
        &\begin{tabular}{@{}c@{}}Wt.\\F1 \end{tabular}
        & \begin{tabular}{@{}c@{}}Time\\NLL \end{tabular}
        & \begin{tabular}{@{}c@{}}Mark\\NLL \end{tabular}
        & \begin{tabular}{@{}c@{}}NLL \end{tabular} 
        & \begin{tabular}{@{}c@{}}NLL/ \\ Time \end{tabular}  
        & \begin{tabular}{@{}c@{}}Micro\\F1 \end{tabular}
        &\begin{tabular}{@{}c@{}}Wt.\\F1 \end{tabular}
        \\
    
    \midrule
        
        \text{CP}				& 57.192 &	17.787 &	74.979 &	0.858 &	53.078 &	45.109 &	348.092	& 100.832 &	448.924 &	0.034 &	30.518 &	28.281\\
		\text{RMTPP}			& 57.301 &	17.117 &	74.418 &	0.851 &	53.273 &	42.261 &	350.847	& 101.094 &	451.941 &	0.034 &	29.951 &	26.958\\
		\text{LNM}				& 56.081 &	16.948 &	73.029 &	0.845 &	57.062 &	52.517 &	341.494	& 98.897 &	440.391 &	0.031 &	44.528 &	42.757\\
		\text{NHP}				& 57.416 &	17.265 &	74.681 &	0.851 &	51.901 &	41.908 &	350.015	& 101.307 &	451.322 &	0.033 &	30.023 &	28.184\\
		\text{SAHP}				& 56.827 &	17.147 &	73.974 &	0.849 &	52.874 &	43.669 &	348.152	& 99.845 &	447.997 &	0.033 &	30.855 &	28.878\\
		\text{THP}				& 56.744 &		\textbf{16.748} &	73.492 &	0.849 &	53.633 &	45.852 &	347.943	& 98.936 &	446.879 &	0.032 &	31.401 &	30.206\\
			
		\textit{\textbf{Prop. RMTPP}}	& 52.795 &	17.018 &	69.813 &	0.806 &	57.063 &	52.621 & 337.743 &	101.983 &	439.726 &	0.031 &	40.425 &	38.153\\
	\textit{	\textbf{Prop. LNM}}		& \textbf{52.566} & {16.942} &	\textbf{69.508} &	\textbf{0.802} &	\textbf{60.372} &	\textbf{58.813} & \textbf{328.301} &	\textbf{98.873} &	\textbf{427.174} &	\textbf{0.028} &	\textbf{56.002} &	\textbf{54.952}\\
	\textit{\textbf{Prop. THP}}		& 52.932 &	17.047 &	69.979 &	0.808 &	56.973 &	52.152 & 334.753 &	100.975 &	435.728 &	0.03 &	42.876 &	42.403\\
    \bottomrule
  \end{tabular}
  
  \label{tab:results_table}
\end{table*}

\begin{table*}[ht]
    \centering
    
    \caption{Robustness of the proposed LNM model with respect to the number of mixture components C.}
    
    \begin{tabular}{c|cc|cc|cc|cc|cc|cc}
    
    \toprule
      & \multicolumn{2}{c|}{Hawkes Ind.} 
      & \multicolumn{2}{c|}{Hawkes Dep. (I)} 
      & \multicolumn{2}{c|}{Hawkes Dep. (II)} 
      & \multicolumn{2}{c|}{MIMIC-II} 
      & \multicolumn{2}{c|}{MOOC} 
      & \multicolumn{2}{c}{Stack Overflow} 
      \\
     
    \midrule
     C  & \begin{tabular}{@{}c@{}}Time\\NLL \end{tabular}  
        & \begin{tabular}{@{}c@{}}Mark\\NLL \end{tabular} 
        & \begin{tabular}{@{}c@{}}Time\\NLL \end{tabular}  
        & \begin{tabular}{@{}c@{}}Mark\\NLL \end{tabular}
        & \begin{tabular}{@{}c@{}}Time\\NLL \end{tabular}  
        & \begin{tabular}{@{}c@{}}Mark\\NLL \end{tabular} 
        & \begin{tabular}{@{}c@{}}Time\\NLL \end{tabular}  
        & \begin{tabular}{@{}c@{}}Mark\\NLL \end{tabular} 
        & \begin{tabular}{@{}c@{}}Time\\NLL \end{tabular}  
        & \begin{tabular}{@{}c@{}}Mark\\NLL \end{tabular}
        & \begin{tabular}{@{}c@{}}Time\\NLL \end{tabular}  
        & \begin{tabular}{@{}c@{}}Mark\\NLL \end{tabular}
        \\
    \midrule
    
    1   & 48.654  & 11.567 & 54.455 & 16.939 & -220.702 & 628.711 & -11.844 & 5.926 & 349.01  & 96.269 & 211.657 & 108.403  \\ 
    2   & 47.495  & 11.568 & 53.595 & 16.938 & -237.75  & 628.706 & -19.478 & 5.757 & 334.442 & 95.438 & 204.965 & 108.204 \\
    4   & 47.201  & 11.572 & 53.127 & 16.939 & -240.703 & 628.709 & -20.591 & 5.977 & 330.769 & 96.093 & 203.775 & 108.445 \\
    8   & 47.202  & 11.567 & 53.132 & 16.938 & -240.931 & 628.712 & -20.475 & 5.938 & 329.379 & 97.765 & 204.543 & 108.553 \\
    16  & 47.165  & 11.57  & 53.125 & 16.939 & -240.931 & 628.706 & -20.377 & 5.838 & 329.485 & 97.881 & 204.784 & 108.816 \\
    32  & 47.189  & 11.573 & 53.106 & 16.939 & -240.937 & 628.72 & -20.635 & 5.741 & 328.766 & 98.325 & 204.442 & 109.767 \\
    64  & 47.176  & 11.574 & 53.122 & 16.938 & -240.925 & 628.746 & -20.464 & 5.903 & 327.396 & 99.283 & 204.567 & 109.86 \\

    \bottomrule
    
    \end{tabular}
  
  \label{tab:mixture_comp_sensitivity}
\end{table*}

\subsection{Baseline Algorithms}

Intensity-based models approximate the integral in Equation \ref{eqn:time_PDF_ground_intensity} using MC estimation. The event history could be encoded either using a recurrent neural network (RNN, LSTM, or GRU) or a self-attention mechanism. We compare against following state-of-the-art models (decoders) on the standard prediction task:

\begin{itemize}
    \item Conditional Poisson \textbf{CP} is a time-independent multi-layer perceptron (MLP) based decoder.
    \item \textbf{RMTPP}: This is an exponential intensity-based decoder agreeing to a Gompertz distribution by \cite{du2016recurrent}. Here, events are encoded using a recurrent neural network. 
    \item \textbf{LNM}: This decoder is intensity-free log-normal mixture model by \cite{shchur2020intensity}. It employs RNN as an event encoder.
    \item \textbf{NHP}: An intensity-based multivariate decoder proposed by \cite{mei2017neuralhawkes}. It uses a continuous-time LSTM encoder for event history encoding. 
    \item \textbf{SAHP}: This model uses a self-attention mechanism for event history encoding operation as discussed in \cite{zhang2020self}.
    \item \textbf{THP}: Transformer based model developed by \cite{zuo2020transformer}. It leverages the self-attention mechanism for long-term event dependency. This model is intensity-based and requires MC approximation in likelihood computation.
\end{itemize}

While SAHP and THP models use attention mechanisms for history encoding, CP, RMTPP, LNM, and NHP use recurrent encoders. Recurrent encoders take $O(N)$ time to encode an event sequence with $N$ events, contrarily, self-attention-based encoders require $O(N^2)$ time. On one hand, CP, RMTPP, LNM, SAHP, and THP are conditionally independent models. On the other hand, NHP is a conditionally dependent model. In the proposed approach, we have two intensity-based models namely, \textit{proposed RMTPP} and \textit{proposed THP}, and one intensity-free model, \textit{proposed LNM}. GRU encodes the event history in \textit{proposed RMTPP} and \textit{proposed LNM}. Our decoders are multivariate, intensity-free mixture (\textit{proposed LNM}) or intensity-based attention models (\textit{proposed THP}) where time distribution is conditioned on all possible marks.

In the following sections, we provide additional technical details of the baselines used.

Conditional Poisson (\textbf{CP}) is a simple time-independent decoder based on multi-layer perceptron (MLP). Let $\pmb{h}_t$ denote the event history vector for all the events occurring before time $t$. CP decodes the history vector $\pmb{h}_t$ into conditional intensity function $\lambda^{*}_{k}(t)$ and cumulative intensity function $\Lambda^{*}_{k}(t)$. Here, subscript $k$ represents mark type. These functions are as follows:

\begin{equation}
    \lambda^{*}_{k}(t) = \text{MLP}(\pmb{h}_t)
    \text{ and }
    \Lambda^{*}_{k}(t) = \text{MLP}(\pmb{h}_t)(t-t_i),
\end{equation}

where, $t_i$ is the arrival time of the event occurring just before time $t$.

\textbf{RMTPP} is an exponential intensity-based unimodal decoder agreeing to a Gompertz distribution and is proposed by \cite{du2016recurrent}. RMTPP is a conditionally independent decoder. Here, the conditional intensity and cumulative intensity are formulated as follows:

\begin{equation}
    \lambda^{*}_{k}(t) = \exp \left( \pmb{W}_1 \pmb{h}_t + w_2 (t-t_i) + \pmb{b}_1 \right )_k
\end{equation}

\begin{equation}
    \Lambda^{*}_{k}(t) = \frac{1}{w_2} [  \exp \left ( \pmb{W}_1 \pmb{h}_t + \pmb{b}_1  \right ) - \exp \left( \pmb{W}_1 \pmb{h}_t + w_2 (t-t_i) + \pmb{b}_1 \right )]_k,
\end{equation}

where, $\pmb{W}_1$, $w_2$, and $\pmb{b}_1$ are learnable parameters of the neural network and 
$\pmb{W}_1 \in \mathbb{R}^{\vert \pmb{h}_t \vert \times K}$, $w_2 \in \mathbb{R} $ and $\pmb{b}_1 \in \mathbb{R}^K$. Note that, $K$ represents the total number of marks and $k$ represents the mark type. 

\textbf{LNM} is an intensity-free log-normal mixture decoder proposed by \cite{shchur2020intensity}. LNM is a conditionally independent decoder that models PDF of inter-event time as follows:

\begin{equation}
    f(\tau \vert \pmb{w}, \pmb{\mu}, \pmb{s}) = \sum_{c=1}^{C} w_{c} \frac{1}{\tau s_{c} \sqrt{2 \pi}} \exp \left(  -\frac{(\log {\tau} - \mu_{c})^{2}}{2s_{c}^2} \right) , 
\end{equation}

where,  $\pmb{w}$ are the mixture weights, $\pmb{\mu}$ are the mixture means and $\pmb{s}$ are the mixture standard deviations. Here, number of mixture component are represented by $c \in \{1,2, \dots, C \} $. The parameters $\pmb{w}, \pmb{\mu} \text{ and } \pmb{s}$ are estimated from $\pmb{h}$ as follows: 

\begin{equation}
    \pmb{w} = \text{softmax} (\pmb{W}_{\pmb{w}} \pmb{h} + \pmb{b}_{\pmb{w}})
\end{equation}
\begin{equation}
    \pmb{\mu} = \exp (\pmb{W}_{\pmb{\mu}} \pmb{h} + \pmb{b}_{\pmb{\mu}})
    \text{ and }
    \pmb{s} = \pmb{W}_{\pmb{s}} \pmb{h} + \pmb{b}_{\pmb{s}},
\end{equation}    
% \begin{equation}
%     \pmb{s} = \pmb{W}_{\pmb{s}} \pmb{h} + \pmb{b}_{\pmb{s}},
% \end{equation}

where, $ \{ \pmb{W}_{\pmb{w}}, \pmb{W}_{\pmb{\mu}}, \pmb{W}_{\pmb{s}}, \pmb{b}_{\pmb{w}}, \pmb{b}_{\pmb{\mu}}, \pmb{b}_{\pmb{s}} \} $ are the learnable parameters of the neural network.
Note that the LNM model does not condition time distribution on marks and shares the same drawbacks as that of conditionally independent models.

For \textbf{NHP, SAHP, and THP} models, we use the parametrized intensity functions specified in the papers \cite{mei2017neuralhawkes, zhang2020self, zuo2020transformer} respectively. We condition these formulation on marks to obtain conditionally dependent TPP as indicated in Equation \ref{eqn:time_PDF_for_mark}.

\subsection{Evaluation Protocols}
To quantify the predictive performance of TPP models, we use the NLL score metric as shown in Equation \ref{eqn:NLL_computation}. Different event sequences could be defined over different time intervals, therefore, we report NLL normalized by time (NLL/time) score. Additionally, as datasets used are multi-class with class imbalance, we report micro F1 score ( accuracy) and weighted F1 score for marks. Ideally, a model should perform equally well on all metrics.

\subsection{Training and Results}
Our experimentation code and datasets are available on GitHub\footnote{\url{https://github.com/waghmaregovind/joint\_tpp}}.
For all datasets, we use $60 \% $ of the sequences for training, $20 \% $ for validation and rest $ 20 \% $ for test. We train the proposed model by minimizing the NLL score (Equation \ref{eqn:NLL_computation}). For a fair comparison, we try out different hyperparameter configurations on the validation split. Using the best set of hyperparameters, we evaluate performance on the test split. The train set size, validation set size, and test set size along with the best set of hyperparameters for each dataset are given in Table \ref{tab:parameters_of_datasets}. Each dataset is defined on a different time scale. For example, start time and end time in the Stack Overflow dataset are in the order of $1e9$. Thus, for numerical stability, many methods scale the time values with the appropriate time scale. As different event sequences have different lengths, we employ batch-level padding on arrival times and event marks to match the batch dimensions. We use zeros as padding values. We minimize the NLL in the training using Adam optimizer \cite{adam}. The learning rate used for all experiments is $1e-3$ with Adam optimizer regularization decay of $1e-5$. We use early stopping in the training with the patience of $50$. We see the performance of the model on the validation set and choose the best model. Finally, we report metrics on the test set.

\begin{figure}[ht]
  \centering
  \includegraphics[width=\linewidth]{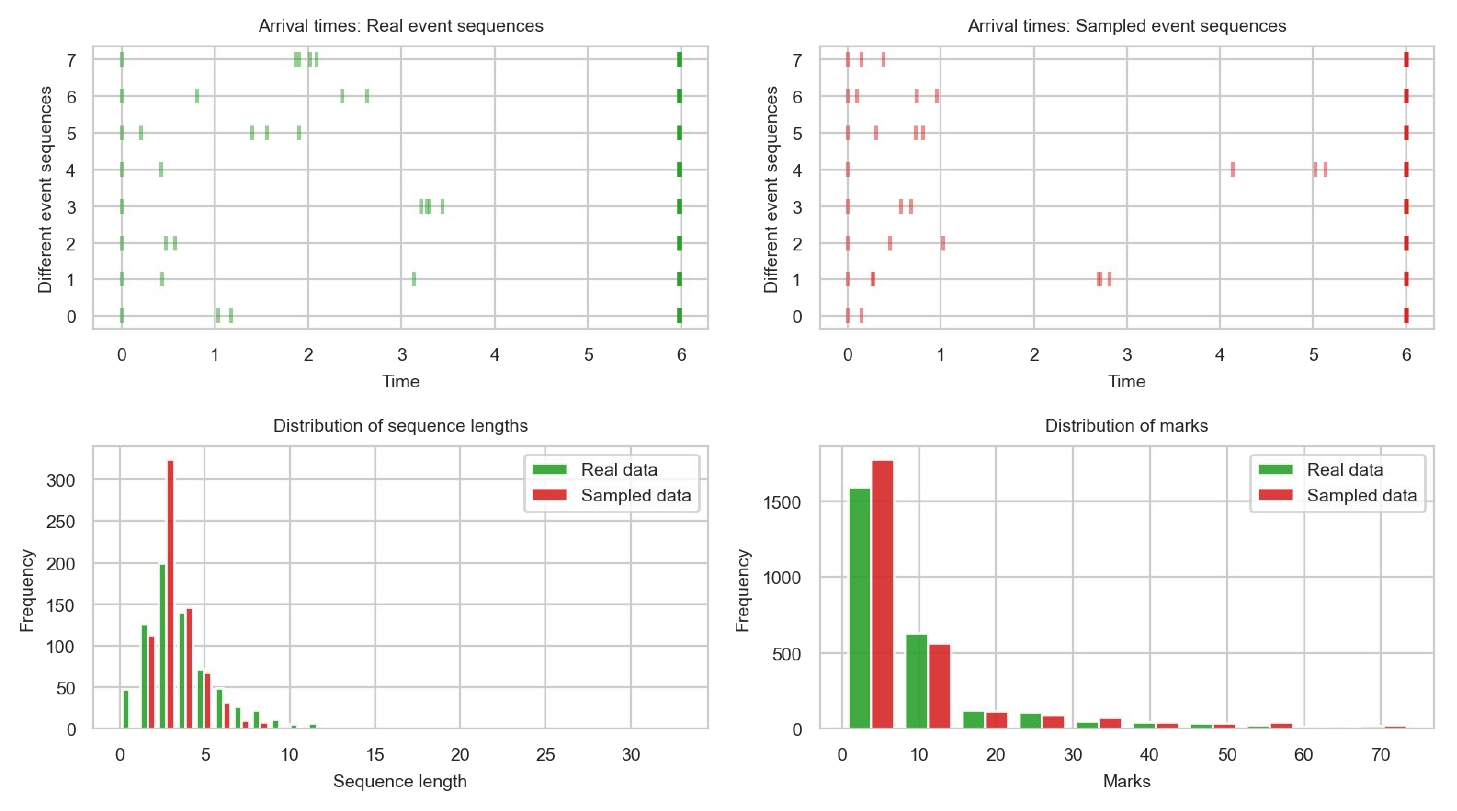}
  \caption{Sampling statistics for MIMIC-II dataset.}
  \label{fig:mimic_ii}
\end{figure}

\begin{figure}[t]
  \centering
  \includegraphics[width=\linewidth]{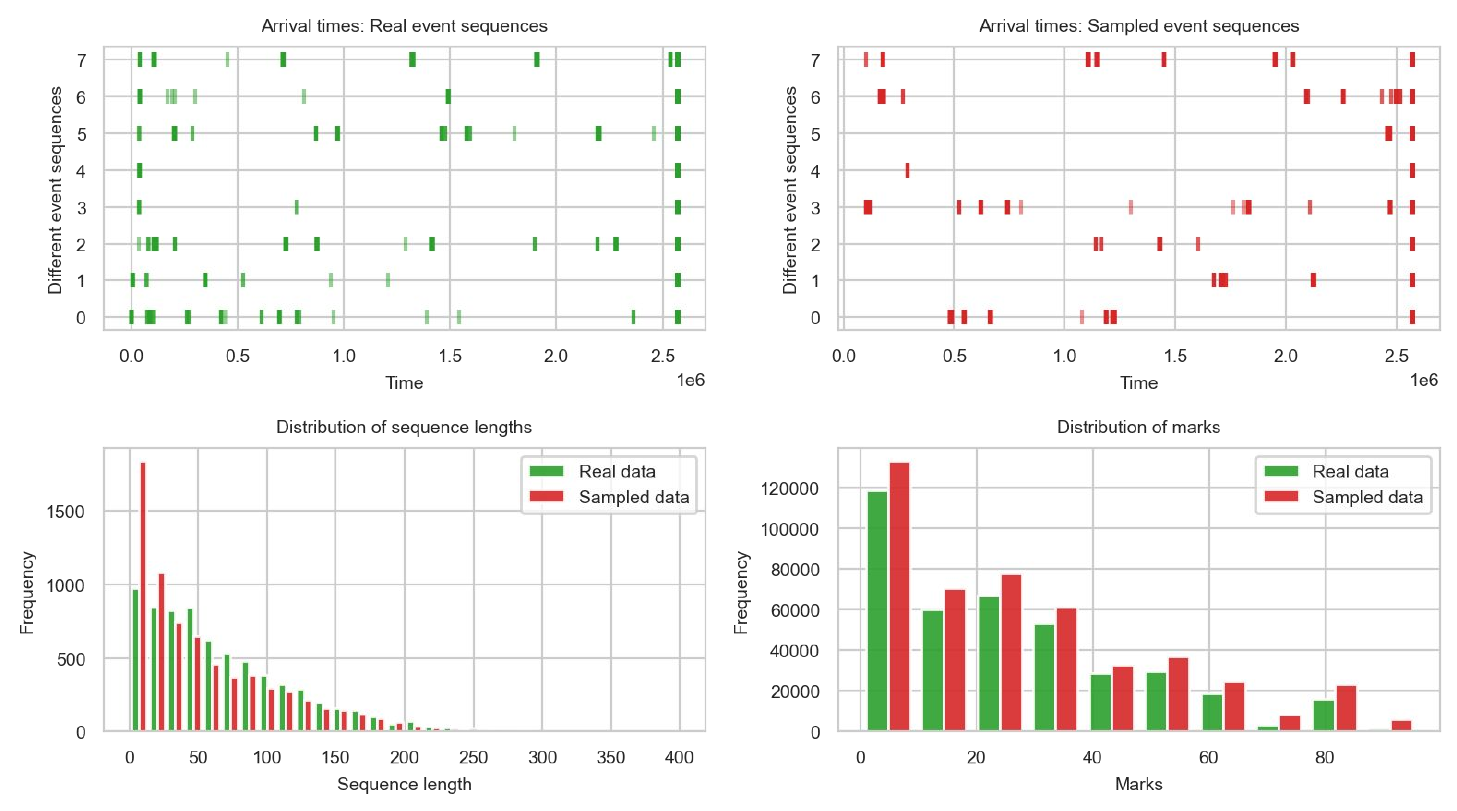}
  \caption{Sampling statistics for MOOC dataset.}
  \label{fig:mooc}
\end{figure}

\begin{figure}[t]
  \centering
  \includegraphics[width=\linewidth]{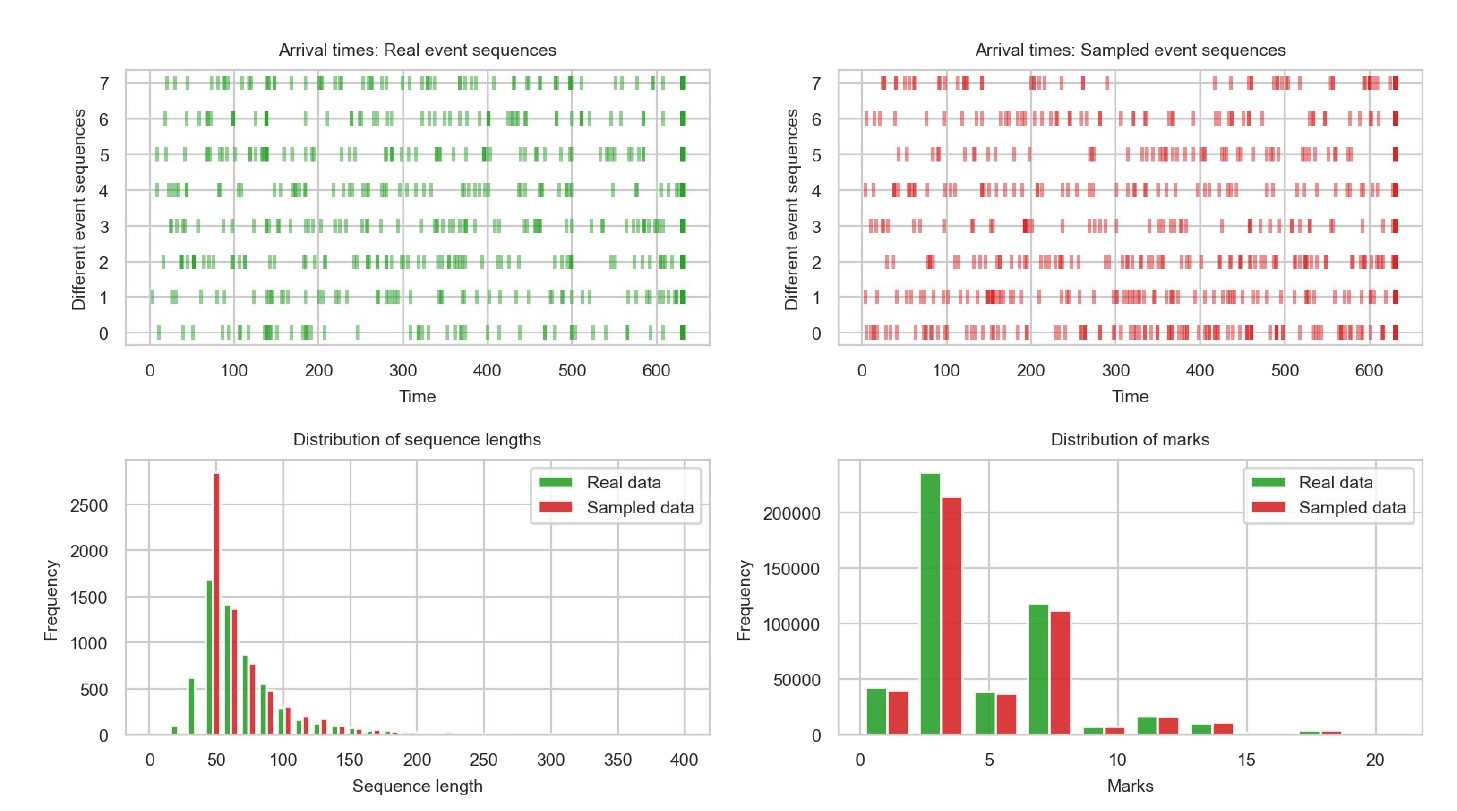}
  \caption{Sampling statistics for Stack Overflow dataset.}
  \label{fig:so}
\end{figure}

The training procedure for the proposed model involves three key steps as shown in Figure \ref{fig:architecture}. These are input representation, event history encoding, and distribution modeling. In the first step, arrival time is converted into inter-event time. The categorical marks are converted into fixed embedding through the mark embedding layer. As different datasets have a different number of marks, we adjust the size of mark embedding accordingly. In the second step, the input representation obtained for all $i-1$ events $(\pmb{y}_{1}, \pmb{y}_{2}, \dots, \pmb{y}_{i-1})$ is passed through RNN to obtain fixed dimensional history vector $\pmb{h}_{i}$ for the $i^{th}$ event. The dimension of the mark embedding and history vector is shown in Table \ref{tab:parameters_of_datasets} as history vector size. Using this history vector $\pmb{h}_{i}$ we model the distribution of inter-event time for all mark types in the final step. 

The predictive performance of the proposed models is shown in Table \ref{tab:results_table}, \ref{tab:results_table_2}, and \ref{tab:results_table_3}. We also provide a breakdown of NLL score into time NLL and mark NLL in the same table to quantify the inter-dependency of time and marks. As emphasized before, 
\textit{A marked TPP model is considered better if it performs well on all the metrics.} The proposed conditionally dependent models show better predictive performance compared to conditionally independent models. All conditionally independent models show similar predictive performance on marks. It is mainly due to the structural design limitation of conditionally independent models. The \textit{proposed LNM} (conditionally dependent) decoder and the LNM decoder are mixture models. Mixture models have universal approximation property to approximate any multimodal distribution \cite{shchur2020intensity}. Due to independence, the LNM mixture model performs poorly compared to conditionally dependent models. In conditionally dependent models, \textit{proposed LNM} model shows superior performance on nearly all the metrics. \textit{Proposed RMTPP} and \textit{proposed THP} models use multivariate intensity-based formulation shown in the Equation \ref{eqn:time_PDF_for_mark}. The likelihood in the training objective does not have a closed-form and requires MC estimates. MC approximation for Equation \ref{eqn:time_PDF_for_mark} is slower and less accurate. Hence, the approximation involved in the likelihood computation is a bottleneck for the predictive performance of the TPPs. As \textit{proposed LNM} model is a conditionally dependent mixture model, we evaluate the likelihood in closed-form. It makes the \textit{proposed LNM} model more flexible and accurate than other conditionally dependent models, as observed in Table \ref{tab:results_table}, \ref{tab:results_table_2} and \ref{tab:results_table_3}.

\begin{table*}[t]
    \centering
    
    \caption{Predictive performance of marked TPP models on synthetic datasets Hawkes independent and Hawkes dependent II. Bold numbers indicate the best performance. Prop. stands for Proposed.}
    
    \begin{tabular}{c|cccccc|cccccc}
  
    \toprule
      
        & \multicolumn{6}{c|}{\textbf{Hawkes Independent}} & \multicolumn{6}{c}{\textbf{Hawkes Dependent II}} \\ 
        \cmidrule(r){2-7} 
        \cmidrule(r){8-13} 
        Model 
        & \begin{tabular}{@{}c@{}}Time\\NLL \end{tabular}
        & \begin{tabular}{@{}c@{}}Mark\\NLL \end{tabular}
        & \begin{tabular}{@{}c@{}}NLL \end{tabular} 
        & \begin{tabular}{@{}c@{}}NLL/ \\ Time \end{tabular}  
        & \begin{tabular}{@{}c@{}}Micro\\F1 \end{tabular}
        &\begin{tabular}{@{}c@{}}Wt.\\F1 \end{tabular}
        & \begin{tabular}{@{}c@{}}Time\\NLL \end{tabular}
        & \begin{tabular}{@{}c@{}}Mark\\NLL \end{tabular}
        & \begin{tabular}{@{}c@{}}NLL \end{tabular} 
        & \begin{tabular}{@{}c@{}}NLL/ \\ Time \end{tabular}  
        & \begin{tabular}{@{}c@{}}Micro\\F1 \end{tabular}
        &\begin{tabular}{@{}c@{}}Wt.\\F1 \end{tabular}
        \\
    
    \midrule
        
        \text{CP}				& 51.304 &	12.058 &	63.362 &	0.726 &	62.271 &	48.474 & -227.08 &		647.219 &	420.139 &	4.22 &	31.51 &	21.269 \\
		\text{RMTPP}			& 50.625 &	11.629 &	62.254 &	0.723 &	63.261 &	48.783 & -239.102 &	646.725 &	407.623 &	4.108 &	32.429 &	22.691\\
		\text{LNM}				& 50.073 &	11.571 &	61.644 &	0.716 &	67.081 &	56.884 & -234.089 &	628.723 &	394.634 &	3.978 &	32.87 &	24.702\\
		\text{NHP}				& 50.598 &	12.66  &	63.258 &	0.728 &	61.927 &	48.137 & -239.38 &		645.472 &	406.092 &	4.143 &	32.023 &	22.539\\
		\text{SAHP}			    & 50.461 &	12.473 &	62.934 &	0.724 &	62.398 &	48.528 & -236.875 &	642.701 &	405.826 &	4.185 &	32.028 &	22.866\\
		\text{THP}				& 50.584 &	12.31  &	62.894 &	0.723 &	63.504 &	48.328 & -238.276 &	639.813 &	401.537 &	4.24 &	32.146 &	22.831\\
								
		\textit{\textbf{Prop. RMTPP}}	& 47.129 &	11.72  &	58.849 &	0.681 &	66.99 &		57.169  & \textbf{-249.928} &	650.631 &	400.703 &	4.038 &	32.742 &	24.315\\
		
		\textit{\textbf{Prop. LNM}}	& \textbf{46.676} &	\textbf{11.571} &	\textbf{58.247} &	\textbf{0.675} &	\textbf{70.802} &	\textbf{65.248} & -240.915 &	\textbf{628.719} &	\textbf{387.804} &	\textbf{3.909} &	\textbf{33.175} &	\textbf{26.256}\\
		
		\textit{\textbf{Prop. THP}}	& 47.061 &	11.691 &	58.752 &	0.68  &	67.128 &	56.885 & -245.588 &	639.305 &	393.717 &	3.955 &	32.721 &	24.367\\
		
    \bottomrule
  \end{tabular}
  
  \label{tab:results_table_2}
\end{table*}

\begin{table*}[t]
    \centering
    
    \caption{Predictive performance of marked TPP models on real datasets Stack Overflow and MIMIC-II. Bold numbers indicate the best performance. Prop. stands for Proposed.}
    
    \begin{tabular}{c|cccccc|cccccc}
  
    \toprule
      
        & \multicolumn{6}{c|}{\textbf{Stack Overflow}} & \multicolumn{6}{c}{\textbf{MIMIC-II}} \\ 
        \cmidrule(r){2-7} 
        \cmidrule(r){8-13} 
        Model 
        & \begin{tabular}{@{}c@{}}Time\\NLL \end{tabular}
        & \begin{tabular}{@{}c@{}}Mark\\NLL \end{tabular}
        & \begin{tabular}{@{}c@{}}NLL \end{tabular} 
        & \begin{tabular}{@{}c@{}}NLL/ \\ Time \end{tabular}  
        & \begin{tabular}{@{}c@{}}Micro\\F1 \end{tabular}
        &\begin{tabular}{@{}c@{}}Wt.\\F1 \end{tabular}
        & \begin{tabular}{@{}c@{}}Time\\NLL \end{tabular}
        & \begin{tabular}{@{}c@{}}Mark\\NLL \end{tabular}
        & \begin{tabular}{@{}c@{}}NLL \end{tabular} 
        & \begin{tabular}{@{}c@{}}NLL/ \\ Time \end{tabular}  
        & \begin{tabular}{@{}c@{}}Micro\\F1 \end{tabular}
        &\begin{tabular}{@{}c@{}}Wt.\\F1 \end{tabular}
        \\
    
    \midrule
        
        \text{CP}				& 218.612 &	118.947 &	337.559 &	0.576 &	43.656 &	30.179 & -18.377 &	5.948 &	-12.429 &	-25.497 &	56.461 &	53.842\\
		\text{RMTPP}			& 224.172 &	117.659 &	341.831 &	0.581 &	43.842 &	28.241 & -18.826 &	5.952 &	-12.874 &	-27.595 &	35.126 &	24.787\\
		\text{LNM}				& 208.131 &	111.137 &	319.268 &	0.534 &	46.451 &	32.925 & -18.703 &	5.927 &	-12.776 &	-26.725 &	66.208 &	63.668\\
		\text{NHP}				& 217.685 &	114.887 &	332.572 &	0.561 &	44.439 &	30.203 & -17.53 &	6.595 &	-10.935 &	-26.332 &	58.274 &	52.125\\
		\text{SAHP}				& 216.995 &	112.686 &	329.681 &	0.557 &	45.139 &	30.892 & -18.384 &	7.089 &	-11.295 &	-27.819 &	59.492 &	53.502\\
		\text{THP}				& 215.093 &	112.544 &	327.637 &	0.551 &	45.459 &	31.861 & -19.73 &	6.511 &	-13.219 &	-28.121 &	60.724 &	54.146\\
								
	\textit{	\textbf{Prop. RMTPP}}	& 220.284 &	119.779 &	340.063 &	0.575 &	45.271 &	29.68 & -21.883 &	5.874 &	-16.009 &	-34.318 &	33.998 &	23.836\\
		\textit{\textbf{Prop. LNM}}	& \textbf{204.344} &	\textbf{110.147} &	\textbf{314.491} &	\textbf{0.531} &	\textbf{47.885} &	\textbf{34.364} & -21.761 &	\textbf{5.849} &	-15.912 &	-33.548 &	\textbf{66.61}  &	\textbf{63.725} \\
		\textit{\textbf{Prop. THP}}	& 211.755 &	113.103 &	324.858 &	0.548 &	46.897 &	33.294 & \textbf{-22.365} &	6.013 &	\textbf{-16.352} &	\textbf{-34.893} &	59.528 &	53.221 \\
       
    \bottomrule
  \end{tabular}
  
  \label{tab:results_table_3}
\end{table*}

Average event sequence length, number of marks, and mark class distribution play a crucial role in the predictive performance of the marked TPP models (see Table \ref{tab:parameters_of_datasets} for statistics). For MIMIC-II, the average sequence length is four. Thus, all models show high variation in the metrics on different data splits. RMTPP performs competitively on simple datasets like Hawkes Ind. and Hawkes Dep. (I) but fails to perform on a dataset with a larger number of marks and longer event sequences. In Table \ref{tab:results_table}, we closely observe impact made by our multivariate TPP model on NLL score. We observe significant improvement in time NLL score as time distribution is conditioned on each mark. Improvement in the time NLL improves marker classification metrics. For conditionally independent models, mark class is inferred as per the Equation \ref{eqn:mark_cond_independent} and for conditionally dependent models marks class is inferred using Equation \ref{eqn:mark_cond_dependent}. MOOC dataset contains interactions of learners with the online courses. Here, the event sequence represents the time-evolving course journey of the learner. Marks represent different activities performed towards course completion. It contains entangled time and marks, and conditionally independent models fail to capture this relationship. The number of marks in the MOOC dataset is $97$. Thus, the intensity-based model numerically approximates $97$ such function using MC estimates. On MOOC, the \textit{proposed LNM}, conditionally dependent mixture model shows boost of $11.5 \% $ on micro F1 score and $12.2 \% $ on weighted F1 score in mark prediction compared to next best model (refer Table \ref{tab:results_table}). It is mainly due to the intensity-free modeling of inter-event time PDF and multivariate formulation. The proposed models consistently outperform other baselines in time and marker prediction tasks on all datasets.

In the \textit{proposed LNM} model, for all datasets, we have used the number of mixture components as $64$. This value is suggested by \cite{shchur2020intensity}, which is equivalent to a number of parameters in the single-layer model proposed by \cite{omi_FNN}. We also provide the sensitivity of NLL metrics with respect to the number of mixture components, C, in Table \ref{tab:mixture_comp_sensitivity}. Empirically, the proposed mixture model is robust to the different values of C. For the \textit{proposed LNM} model, the NLL function does not contain any integration term as inter-event time PDF is modeled using mixture models. Therefore, we leverage mixture models to estimate likelihood in closed-form. In closed-form sampling, we first sample the categorical mark distribution. Using this sampled mark of type $m_i=k$, we select the time PDF $f_{k}(\tau \vert \pmb{w}_{k}, \pmb{\mu}_{k}, \pmb{s}_{k})$. Further, we sample from this PDF to get the next inter-event time $\tau_i$ of mark type $k$. To evaluate sampled event sequences qualitatively, we plot arrival times, distribution of event sequence lengths, and distribution of marks for each dataset. Sampling analysis for real-world datasets is shown in Figures \ref{fig:mimic_ii}, \ref{fig:mooc}, and \ref{fig:so}. The total NLL score consists of the time NLL component of continuous inter-event time and the mark NLL component of categorical marks. Both these components play a key role in model training and influence future predictions. In Table \ref{tab:results_table}, we provide a breakdown of the NLL score for all the models. The proposed conditionally dependent models show better time NLL and the mark NLL value due to multivariate modeling.

\section{Limitations and Conclusion}
Conditionally dependent models use multivariate formulation to condition inter-event time distribution on the set of categorical marks. If the number of marks $K$ is extremely large, mark prediction becomes an extreme class classification problem. To address this, \cite{Guo2018INITIATORNE, mei-2020-nce} have proposed noise-contrastive-estimation-based models. 

In this work, we discuss the adverse effect of the independence assumption between time and mark on the predictive performance of the marked TPPs. We address this structural shortcoming by proposing a conditionally dependent multivariate TPP model under both intensity-based and intensity-free settings. The \textit{proposed LNM} architecture overcomes the drawbacks of an intensity-based conditionally dependent model and poses desired properties like closed-form likelihood, and closed-form sampling. Multiple evaluation metrics on diverse datasets highlight the impact of our work against state-of-the-art conditionally dependent and independent marked TPP models.

\bibliographystyle{ACM-Reference-Format}
\balance
\bibliography{cikm}

\end{document}